\documentclass[journal]{IEEEtran}

\usepackage{graphicx}
\usepackage{multirow}
\usepackage{epstopdf}

\usepackage{amsmath,amssymb}

\usepackage{paralist}
\usepackage{algorithm}
\usepackage{algorithmicx}
\usepackage{algpseudocode}
\usepackage{array}
\usepackage{comment}

\usepackage{caption}
\usepackage{subcaption}

\begin{document}

\title{Learning to detect video events from zero or very few video examples}

\author{Christos~Tzelepis,
	Damianos Galanopoulos,
        Vasileios~Mezaris,
        and~Ioannis~Patras
\thanks{C. Tzelepis is with the Information Technologies Institute/Centre for Research and 
        Technology Hellas (CERTH), Thermi 57001, Greece, and also with the School of 
        Electronic Engineering and Computer Science, Queen Mary University of London, 
        London E1 4NS, U.K. (email: tzelepis@iti.gr).}
\thanks{D. Galanopoulos is with the Information Technologies Institute/Centre for Research and 
        Technology Hellas (CERTH), Thermi 57001, Greece (email: dgalanop@iti.gr).}
\thanks{V. Mezaris is with the Information Technologies Institute/Centre for Research and 
        Technology Hellas (CERTH), Thermi 57001, Greece (email: bmezaris@iti.gr).}
\thanks{I. Patras is with the School of Electronic Engineering and Computer Science, 
        Queen Mary University of London, London E1 4NS, U.K. (e-mail: i.patras@qmul.ac.uk).}
        }

\maketitle

\begin{abstract}
      In this work we deal with the problem of high-level event detection in video. Specifically, we study the challenging problems of i) learning to detect video events from solely a textual description of the event, without using any positive video examples, and ii) additionally exploiting very few positive training samples together with a small number of ``related'' videos. For learning only from an event's textual description, we first identify a general learning framework and then study the impact of different design choices for various stages of this framework. For additionally learning from example videos, when true positive training samples are scarce, we employ an extension of the Support Vector Machine that allows us to exploit ``related'' event videos by automatically introducing different weights for subsets of the videos in the overall training set. Experimental evaluations performed on the large-scale TRECVID MED 2014 video dataset provide insight on the effectiveness of the proposed methods.
\end{abstract}


\IEEEpeerreviewmaketitle


\section{Introduction}\label{sec:intro}

      \IEEEPARstart{H}{igh}-level (or complex) video event detection is the problem of finding, within a set of videos, which of them depict a given event. Typically, an event is defined as an interaction among humans or between humans and physical objects \cite{Jiang13}. Some examples of complex events are those defined in the Multimedia Event Detection (MED) task of the TRECVID benchmarking activity \cite{2013trecvidover,2014trecvidover}. For instance, in MED $2014$ \cite{2013trecvidover}, the defined complex events include \textit{Attempting a bike trick}, \textit{Cleaning an appliance}, or \textit{Beekeeping}, to name a few. 
      
      The detection of such events in video has recently drawn significant attention in a wide range of applications, including video annotation and retrieval \cite{Jiang13}, video organization and summarization \cite{zhang2002event}, or surveillance applications \cite{oh2011large}. In \cite{Brown05}, Brown studies how high-level events play a substantial role in the mechanism of structuring memories and recalling past experiences. This leads to the expectation that event-based organization of video content can significantly contribute to bridging the existing semantic gap between human and machine understanding of multimedia content.
      
      There are several challenges associated with building an effective detector of video events. One of them is finding a video representation that reduces the gap between the traditional low-level audio-visual features that can be extracted from the video and the semantic-level actors and elementary actions that are by the definition the constituent parts of an event. In this direction, several works have shown the importance of using simpler visual concepts as a stepping stone for detecting complex events (e.g. \cite{gkalelis2013video, gkalelis2014video}). Another major challenge is to learn an association between the chosen video representation and the event or events of interest; for this, supervised machine learning methods are typically employed, together with suitably annotated training video corpora. While developing efficient and effective machine learning algorithms is a challenge in its own right, finding a sufficient number of videos that depict the event so as to use them as positive training samples for training any machine learning method is also not an easy feat. In fact, video event detection is even more challenging when the available positive training samples are limited, or even non-existent; that is, when one needs to train an event detector using only textual information that a human can provide about the event of interest.
      
      In this work we study the problems of i) learning an event detector solely from a textual description of the event, without using any positive video examples, and ii) learning from very few positive training samples together with a small number of ``related'' videos. The paper is organized as follows. In Section \ref{sec:rel_work}, related work in video event detection using zero or a few positive examples is reviewed. In Section \ref{sec:textual}, we present and examine different design choices for a framework that learns video event detectors based solely on textual information for training, while in Section \ref{sec:text_fewpos_comb} the combination of the above methods with learning from a few positive examples, as well as from related training examples, is examined. Results of the application of the proposed techniques to the TRECVID MED 2014 dataset are provided in Section \ref{sec:exp_results}. Finally, conclusions are drawn and discussed in Section \ref{sec:conclusion}.


\section{Related Work}\label{sec:rel_work}
      
      There are two broad categories concerning the learning conditions under which a video event detector is trained. That is, training may be done either by using a number of positive and negative training video examples, or using no video examples at all. Within the first category, one can distinguish between i) methods assuming that positive video examples are in abundance, and ii) methods explicitly embracing the fact that the positive samples that are available for training are typically limited, in practice. Regardless of the assumptions on the number of positive training samples, it is assumed that negative samples can be found without much effort, and thus the number of negative video samples is not a restrictive parameter in the process of learning. The above training conditions are typically simulated by the $100$Ex and $10$Ex MED subtasks of TRECVID \cite{2014trecvidover}, where $100$ and $10$ positive video samples are available for training video event detectors, respectively. 
      
      Training based solely on a textual description of each event class is reported in a few works, mostly in the context of the TRECVID MED $0$Ex and Semantic Query (SQ) subtasks \cite{2014trecvidover}, where no positive video examples are provided. Instead, event detectors are trained using textual resources, which typically include the event's title, a short free-form text explanation of what may be depicted in a video that belongs to this event class, as well as brief references to visual and audio cues that are typically expected to be present in 
      such a video (Fig.~\ref{fig:event_kit}).

      \subsection{Learning from zero positive examples}\label{subsec:rel_work_zero}
	    
	    Learning from zero positive examples has recently drawn significant attention in various learning problems, due to its challenging nature and the extensive applicability it has. For instance, due to the rapidly increasing number of images on the Web, extensive research efforts have been devoted in multi-label, zero-example (or few-example) classification in images \cite{mensink2014costa}. Similarly, a method for zero-example classification of fMRI data was proposed in \cite{NIPS2009_3650}. In \cite{elhoseiny2013write}, a method for predicting unseen image classes from a textual description, using knowledge transfer from textual to visual features, was proposed.
	    
	    In the video domain, learning from zero positive examples is investigated primarily in the context of video event detection. In \cite{habibian2014videostory} this problem is addressed by transforming both the event's textual description and the visual content of un-classified videos in a high dimensional concept-based representation, using a large pool of concept detectors; then relevant videos are retrieved by computing the similarities between these representations. Similarly, in \cite{BBNviser14}, each event class title is used as an input query to a text retrieval system, and the most relevant documents are retrieved. The vectorized words of these documents are then projected into the most semantically similar concepts from different modalities, such as ASR (Automatic Speech Recognition), OCR (Optical Character Recognition), and high-level features coming from applying audio-visual and DCNN (Deep Convolutional Neural Networks) concept detectors. Using these concepts, relevant videos are retrieved, and late fusion is used for combining the different ranked lists of videos so as to generate the final event detection results. In \cite{wu2014zero}, multiple low-level representations using both the visual and the audio content of the videos are extracted, along with higher-level semantic features coming from ASR transcripts, OCR, and off-the-shelf video concept detectors. This way, both audio-visual and textual features are expressed in a common high-dimensional concept space, where the computation of similarity is possible. In \cite{habibian2014composite}, logical operators are used to discover different types of composite concepts, which leads to better event detection performance.
	    
	    Moreover, in \cite{jiang2014zero}, a relevance feedback approach is used in order to improve event detection results in the zero-example problem using features computed from several modalities. The main idea is to use the textual information that describes the event class in order to create queries for each modality. Then, the system results in ranked video lists, one per each modality. The top videos from these lists are used as a ``pseudo label'' video set on which a joint model is trained, and a new ranked list is produced and used for creating a new ``pseudo label'' set; this process is iterated a few times.
	 
	    In \cite{jiang2015bridging}, E-Lamp was proposed, which is a zero-example event detection system made of four subsystems. The first one is an off-line indexing component, while the rest of them compose the on-line event search module. In the off-line module, each video is represented with $4043$ visual concepts along with ASR and OCR high-level features. Then, in the on-line search module, the user-specified event description is translated into a set of relevant concepts, called \textit{system query}. This system query is used to retrieve the videos that are most relevant to the event. Finally, a pseudo-relevance feedback approach is exploited in order to improve the results.
	    
	    Our approach goes beyond the classic semantic similarity comparison between a given event title (or other user-specified event cues) and each concept title from a concept pool. We try to enrich each concept by automatically searching in Google or Wikipedia in order to find more information for it; this enables finding semantic similarities between events and concepts more effectively. Exploiting Google Search or Wikipedia is, to the best of our knowledge, novel.

      \subsection{Learning from a few positive examples}\label{subsec:rel_few_ex}
	    
	    A limited number of studies have considered the problem of learning video event detectors from very few (e.g. 10) positive training examples \cite{yuinformedia, SESAME14, CERTH14, tzelepis2013improving}. In \cite{yuinformedia}, visual static (e.g. SIFT \cite{lowe2004distinctive}, Transformed Color Histogram \cite{van2010evaluating}), and motion (e.g. MoSIFT \cite{chen2009mosift}, Improved Dense Trajectories \cite{yuinformedia}) descriptors are used, along with the Fisher Vector encoding scheme \cite{sanchez2013image, perronnin2010improving, chatfield2011devil}. ASR and OCR techniques are also used for exploiting audio and textual information in video streams, as well as audio features and visual features based in DCNNs trained in ImageNet \cite{deng2009imagenet}. DCNN-based features typically comprise one or more of the network's hidden layers \cite{simonyan2014very}, providing high discriminative power \cite{oh2011large}. Based on these features, video event detectors are trained using multiple SVM classifiers and fusion techniques (early and late) for combining different modalities. In SESAME \cite{SESAME14}, the authors also use DCNN classification scores as video features, which are subsequently fed to a kernel SVM  for obtaining event detectors. ASR and low-level audio features are also extracted, and a logistic regression-based fusion technique is used for combining scores from different modalities in order to obtain the final event detectors. In \cite{BBNviser14, Aurora14, CERTH14}, a similar training framework for learning from a few positive samples is used; low-level visual and/or audio features are combined with concept detectors' output scores for obtaining event detectors. Concept detectors are trained either using low-level features and VLAD vectors \cite{jegou2012aggregating}, or DCNN output layers.
	    
	    Furthermore, it is not unusual to have in the training set videos that do not exactly fulfill the requirements to be characterized as true positive examples, but nevertheless are closely related to an event class and can be seen as ``related'' examples of it. This is simulated in the TRECVID MED task \cite{2014trecvidover} by the ``near-miss'' video examples provided for each target event class. Differently from our method, none of the above works takes full advantage of these related videos for learning from few positive samples; instead, the ``related'' samples are either excluded from the training procedure, or they are treated as true positive or true negative instances \cite{douze2014inria}.
	    
	    In our study, we will consider ``related'' training samples as videos that can contribute to event detector training but do not merit being treated as either true positive or true negative instances. To take advantage of them, we employ a Relevance Degree SVM (RDSVM), which can treat these samples as either weighted negatives or weighted positives in conjunction with an automatic weighting selection scheme.


\section{Detecting video events using an event's textual description}\label{sec:textual}

      \subsection{General architecture}\label{subsec:textual_gen_arch}
	    
\begin{figure}
		  \centering
		  \includegraphics[width=9cm]{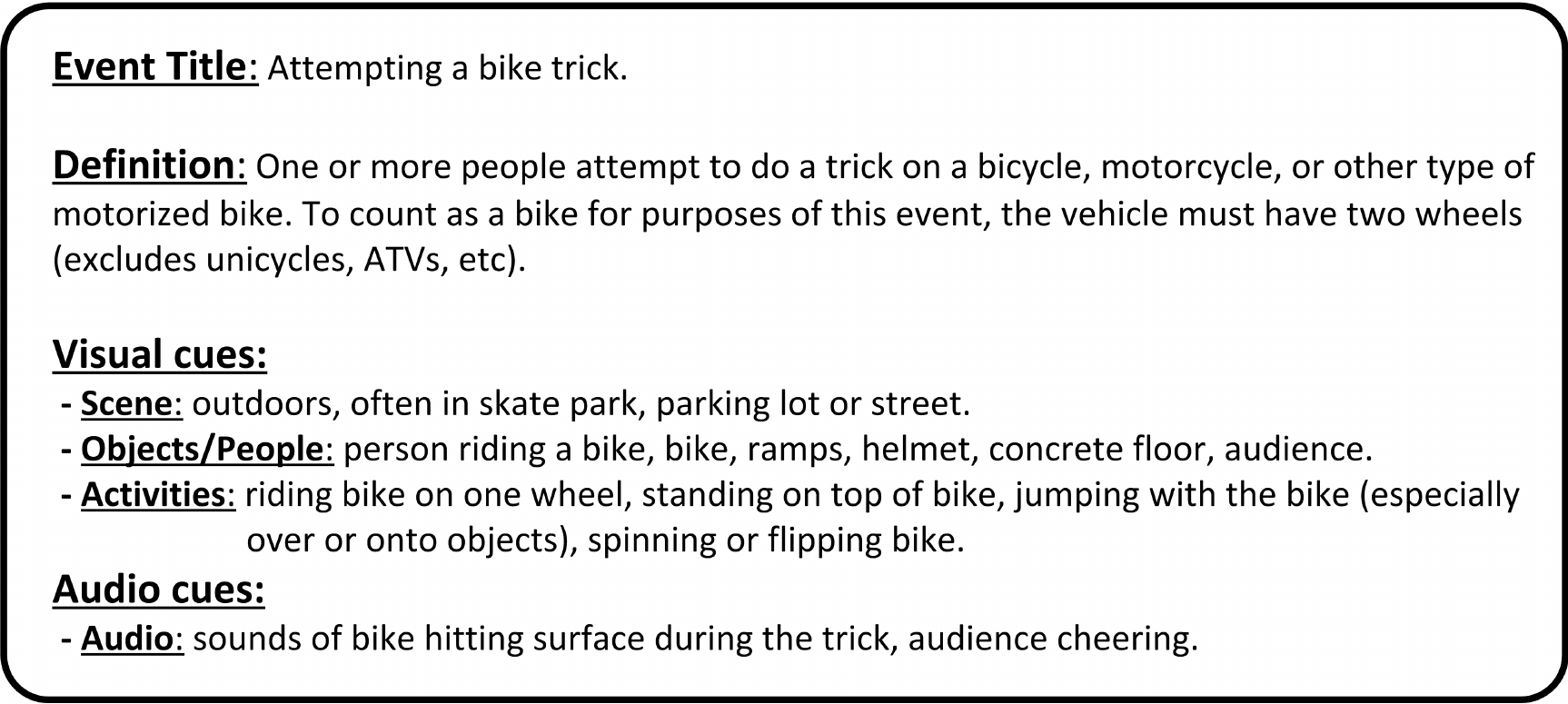}
		  \caption{Textual description of the event class \textit{Attempting a bike trick}.}
		  \label{fig:event_kit}
	    \end{figure}

\begin{figure}
		  \begin{subfigure}[b]{\linewidth}
			\centering
			\includegraphics[width=0.8\textwidth ]{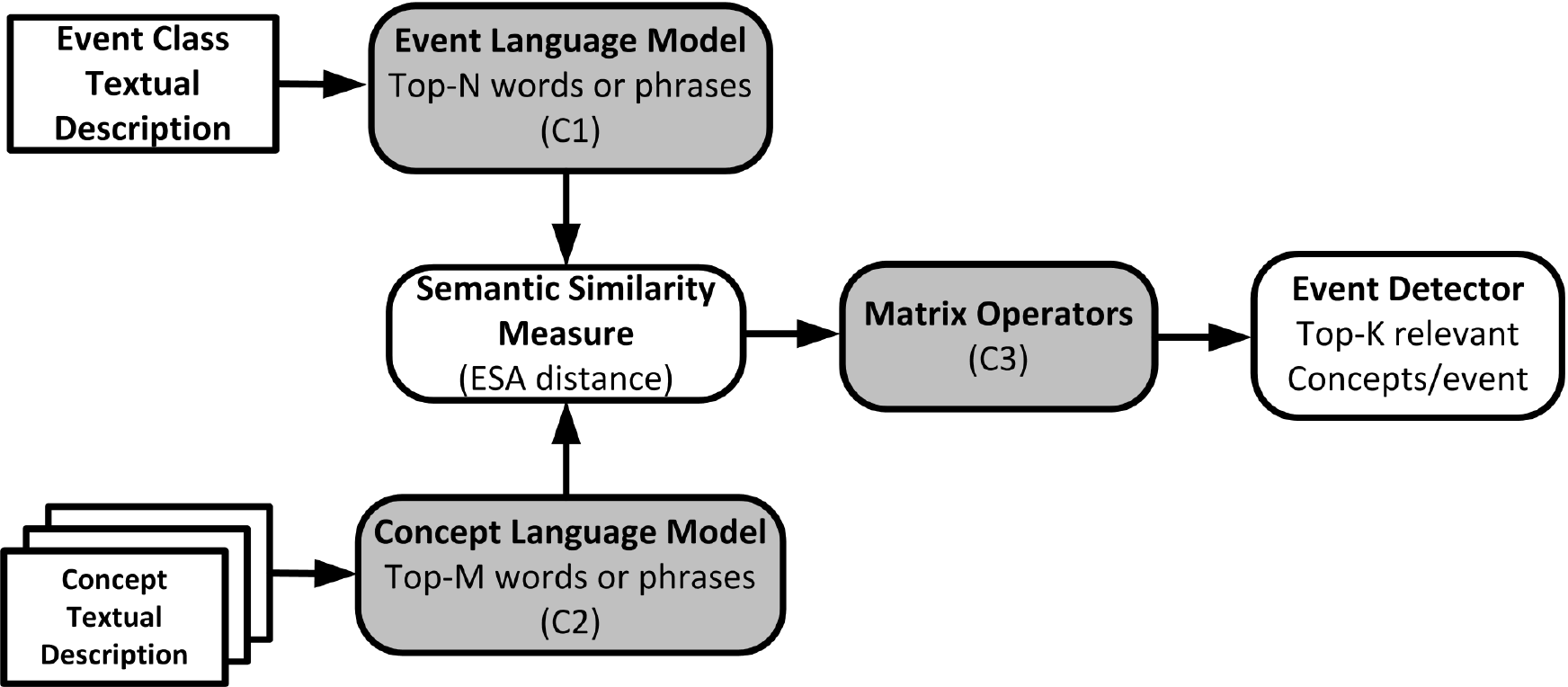}
			\caption{Creation of event detector without positive video samples.}
			\label{fig:ZeroEx_Frameworks_1}
		  \end{subfigure}
		  \\ \vfill
		  \begin{subfigure}[b]{\linewidth}
			\centering
			\includegraphics[width=0.6\textwidth ]{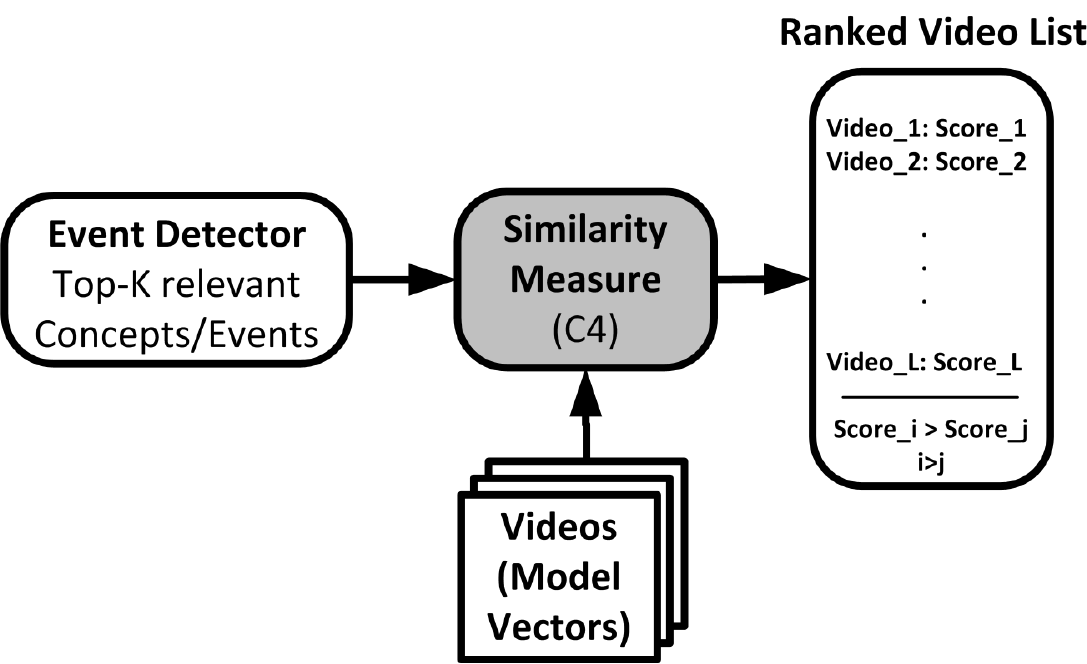}
			\caption{Event detection in a video collection, using the event detector of the above sub-figure.}
			\label{fig:ZeroEx_Frameworks_2}
		  \end{subfigure}
		  \\ \vfill
		  \begin{subfigure}[b]{\linewidth}
			\centering
			\includegraphics[width=0.6\textwidth ]{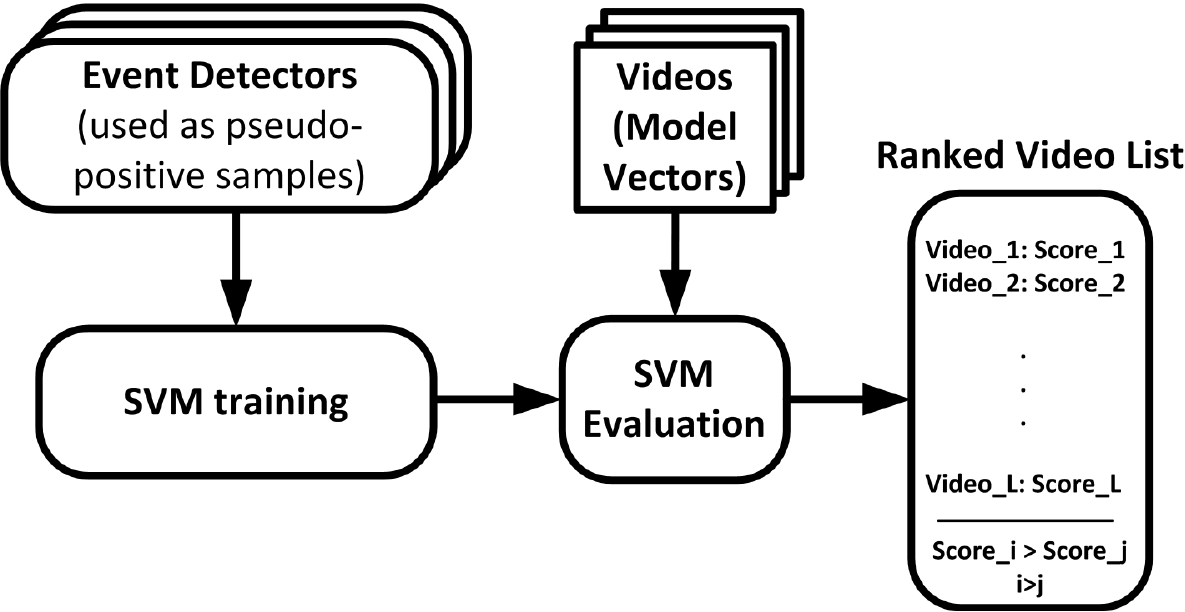}
			\caption{Pseudo-positive sample creation and training.}
			\label{fig:ZeroEx_Frameworks_3}
		  \end{subfigure}
		  \caption{The proposed framework for detecting video events using zero positive examples.}
		  \label{fig:ZeroEx_Frameworks_All}
	    \end{figure}

	    In this section we propose a framework for video event detection without using any visual knowledge about the events. We assume that the only knowledge available, with respect to each event class, is a textual description of it, which consists of a title, a free-form text, and a list of possible visual and audio cues, as in \cite{Younessian,jiang2014zero}. Fig.~\ref{fig:event_kit} shows an example of such a textual description for the event class \textit{Attempting a bike trick}. For linking this textual information with the visual content of the videos that we want to examine, similarly to \cite{wu2014zero,yuinformedia}, we a) use a pool of $N_c$ concepts along with their titles and in some cases a limited number of subtitles (e.g. concept \textit{bicycle-built-for-two} has the subtitles \textit{tandem bicycle} and \textit{tandem}), and b) a pre-trained detector (based on DCNN output scores) for each concept. Fig.~\ref{fig:ZeroEx_Frameworks_All} illustrates the structure of the proposed framework. Shaded blocks indicate processing stages for which different design choices are examined in this work. Each of these stages is discussed below in detail, while Table \ref{tbl:Design_Choices} sums up the different considered design choices.
	    
	    Figure~\ref{fig:ZeroEx_Frameworks_1} shows a first process that receives a textual description of an event class and a list of concepts, and generates an event detector. Given the textual description of the event class, our framework first identifies $N$ words or phrases that most closely relate to the event class; we call this word-set the Event Language Model (ELM). In parallel, for each of the $N_c$ concepts of our concept pool, it similarly identifies $M$ words or phrases; we call this set the Concept Language Model (CLM) of the corresponding concept.

	    Subsequently, for each word in ELM and each word in each one of CLMs we calculate the Explicit Semantic Analysis (ESA) distance \cite{gabrilovich2007computing} between them. For each CLM, the resulting $N\times M$ distance matrix expresses the relation between the given event class and the corresponding concepts. In order to compute a single score expressing this relation, we apply to this matrix different operators, such as various matrix norms or distance measures. Consequently, a score is computed for each pair of ELM and CLM. The $N_c$ considered concepts are ordered according to these scores (in descending order) and the $K$ top concepts along with their scores constitute our event detector. In order to perform event detection in a video collection, we compare this event detector with the output scores of concept detectors applied on each video, using different similarity measures (Fig.~\ref{fig:ZeroEx_Frameworks_2}). Thus, the final output is a ranked list of the most relevant videos. Alternatively, multiple event detectors can be generated using more than one different algorithm variations in each of the shaded blocks of Fig.~\ref{fig:ZeroEx_Frameworks_1} and \ref{fig:ZeroEx_Frameworks_2}, and these can be used as pseudo-positive samples for training an SVM, which can then be applied to the videos so as to generate a ranked list of those depicting the target event (Fig.~\ref{fig:ZeroEx_Frameworks_3}).
	    
      \subsection{Language Models}\label{subsec:Lang_Models}
      
	    We examine the construction of three different types of ELMs, depending on the textual information that they use (Fig.~\ref{fig:event_kit}). The first type of ELM, which will be denoted as ``Title'' hereafter, is based on the automatic extraction of word terms solely from the title of an event class. The second type of ELM, denoted as ``Visual'', is constructed by using only the visual cues provided along with the textual description of each event class. Both title and visual cues consist only of words and single phrases, such as \textit{attempting a bike trick}, \textit{bike}, \textit{riding bike on one wheel}, etc. These words can automatically be extracted from the textual description without any human-expert intervention. Finally, the third type of ELM is obtained by the automatic extraction of words based on the visual and audio cues, as well as on the short free form text of an event class, and it is denoted as ``AudioVisual''.
	    
	    Accordingly, a CLM is constructed for each one of the $N_c$ concepts. We examine the construction of six different types of CLMs, depending on the textual information used for each concept, as well as the weighting technique (e.g. Tf-Idf) adopted for transforming this textual information in a Bag of Words (BoW) \cite{salton1983introduction} representation. As a first approach, the title of a concept, along with any available subtitle, are used as a query to the Google search engine, and the text of the top-$20$ search results per query are retrieved. By applying text cleaning techniques (removing html tags, stop words, etc.) and Bag of Words statistics, we select the most frequently occurring words. These, together with the concept's title and subtitles, constitute the $M$ words of the CLM. As a second approach, the concept title and any subtitle are similarly used as a query in Wikipedia, and the $20$ most relevant articles per query are retrieved. By following the same procedure as above, we end up with the top-$M$ words of the concept's CLM. As a third approach, we use only the title and the subtitles in the CLM. In the three above approaches, the Bag-of-Words (BoW) can be constructed with or without using Tf-Idf weighting \cite{leskovec2014mining}, this resulting in six different types of CLMs.
	    
	    The above types of ELMs and CLMs are introduced as different design choices, as shown in Fig.~\ref{fig:ZeroEx_Frameworks_1}. Table \ref{tbl:Design_Choices} summarizes the specific types of each one of them.

      \subsection{Building an event detector}\label{subsec:build_event_detector}

	    The constructed language models represent the given event class and each of the available concepts as ranked lists of words. Thus, we can calculate the similarities between them by computing the semantic similarity between each word in the ELM and each word belonging to the CLMs. To this end, we use the ESA semantic relatedness measure \cite{carvalho2014easyesa}, which calculates the similarity distance between two terms by computing the cosine similarity between their weighted vectors of Wikipedia articles. In this way, an $N\times M$ matrix with scores for each pair of event-concept is computed. Let $S$ denote the aforementioned similarity matrix; that is, its $(i,j)$-th entry, $s_{i,j}$, denotes the similarity distance between the word $w_i$ in ELM and the word $w_j$ in the respective CLM. Then, we can arrive at a single score expressing the relation between the above ELM and CLM by evaluating one of the matrix operators
	    shown in Table \ref{tbl:Design_Choices} (row C$3$). It is worth noting that $\lambda_{\max}(S^\top S)$ denotes the maximum eigenvalue of the covariance matrix $S^\top S$.
	    	    
	    \begin{table*}[!th]
	    \footnotesize
	    \centering
		\begin{tabular}{|rc|cl|}
		    \hline
		    \multicolumn{2}{|c|}{\textbf{Processing Stage}}                & \multicolumn{2}{c|}{\multirow{2}{*}{\textbf{Examined design choices for this processing stage}}}                                                   \\
		    \multicolumn{2}{|c|}{(see Fig.~\ref{fig:ZeroEx_Frameworks_2})} & \multicolumn{2}{c|}{}                                                                                                                              \\ \hline\hline
		    \multirow{3}{*}{\textbf{C1:}}                                  & \multirow{3}{*}{ELM}                                                                             & \multicolumn{2}{c|}{i) Title}                                                                                                                      \\
						                                   &                                                                                                  & \multicolumn{2}{c|}{ii) Visual only}                                                                                                               \\
						                                   &                                                                                                  & \multicolumn{2}{c|}{iii) AudioVisual}                                                                                                              \\ \hline
		    \multirow{3}{*}{\textbf{C2a:}}                                 & \multirow{3}{*}{\begin{tabular}[c]{@{}c@{}}CLM\\ information\\sources\end{tabular}}                                                                             & \multicolumn{2}{c|}{i) Title}                                                                                                                      \\
						                                   &                                                                                                  & \multicolumn{2}{c|}{ii) Google}                                                                                                                    \\
						                                   &                                                                                                  & \multicolumn{2}{c|}{iii) Wikipedia}                                                                                                                \\ \hline
		    \multirow{2}{*}{\textbf{C2b:}}                                 & \multirow{2}{*}{\begin{tabular}[c]{@{}c@{}} CLM\\weighting\end{tabular}}                                                                      & \multicolumn{2}{c|}{i) Tf-Idf}                                                                                                                     \\
						                                   &                                                                                                  & \multicolumn{2}{c|}{ii) No Tf-Idf}                                                                                                                 \\ \hline
		    \multirow{5}{*}{\textbf{C3:}}                                  & \multirow{5}{*}{\begin{tabular}[c]{@{}c@{}}Matrix\\ Operation\end{tabular}}                      & \multicolumn{1}{l}{i) $\ell_2$ Norm}       & $\|S\|_2 = \sqrt{\lambda_{max}\big(S^\top S\big)}$                                                    \\
						                                   &                                                                                                  & \multicolumn{1}{l}{ii) $\ell_\infty$ Norm} & $\|S\|_\infty = \max_{1\leq i \leq N}\sum_{j=1}^{M}|s_{i,j}|$                                         \\
						                                   &                                                                                                  & \multicolumn{1}{l}{iii) Frobenius Norm}    & $\|S\|_{F} = \Bigg(\sum_{i=1}^{N}\sum_{j=1}^{M} |s_{i,j}|^2\Bigg)^{\frac{1}{2}}$                      \\
						                                   &                                                                                                  & \multicolumn{1}{l}{iv) Maximum entry}      & $\max\big(S\big)$                                                                                     \\
						                                   &                                                                                                  & \multicolumn{1}{l}{v) Hausdorff Distance}  & $D_{\mathcal{H}}(\text{EML}, \text{CML})=\underset{p}{\mathrm{median}}\Big(\max_{p}\|w_i-w_j\| \Big)$ \\ \hline
		    \multirow{5}{*}{\textbf{C4:}}                                  & \multirow{5}{*}{Distance}                                                                        & \multicolumn{2}{c|}{i) Cosine}                                                                                                                     \\
						                                   &                                                                                                  & \multicolumn{2}{c|}{ii) Histogram Intersection}                                                                                                    \\
						                                   &                                                                                                  & \multicolumn{2}{c|}{iii) Kullback}                                                                                                                 \\
						                                   &                                                                                                  & \multicolumn{2}{c|}{iv) $X^2$}                                                                                                                     \\
						                                   &                                                                                                  & \multicolumn{2}{c|}{v) Euclidean}                                                                                                                  \\ \hline
		\end{tabular}
		\caption{Design choices.}
		\label{tbl:Design_Choices}
	    \end{table*}
	    
	    \begin{figure*}
		  \centering
		  \includegraphics[width=\textwidth]{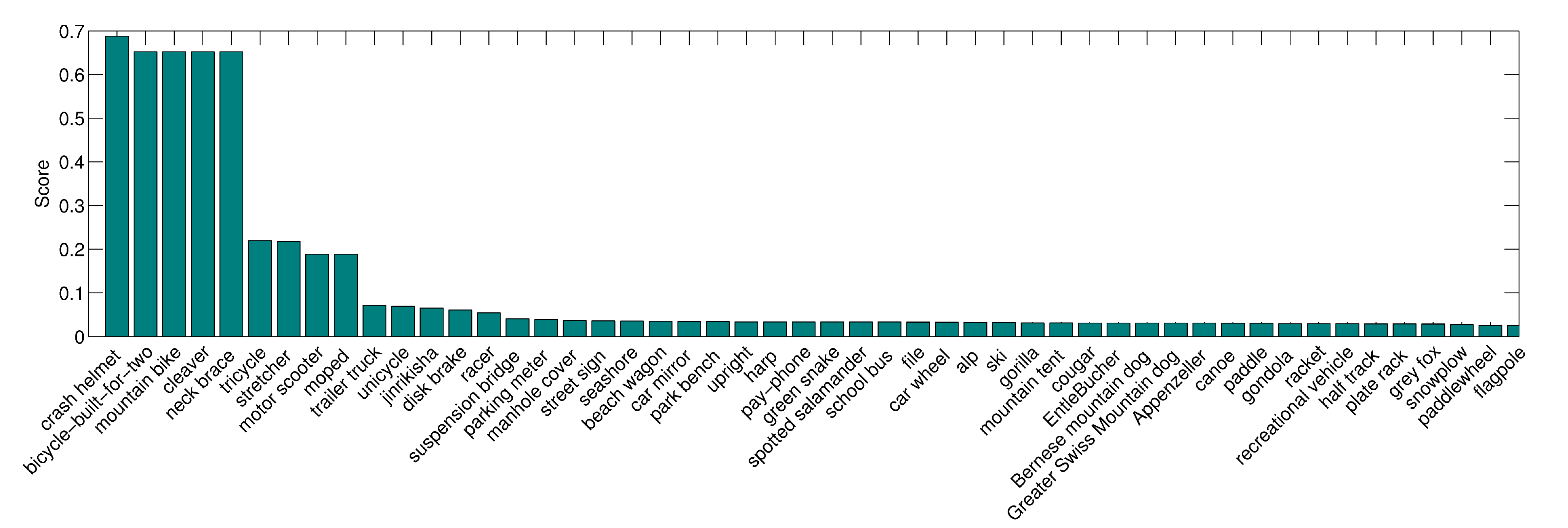}
		  \caption{Example of event detector for the event class \textit{Attempting a bike trick}.}
		  \label{fig:E021_Detector}
	    \end{figure*}
	    
	    Following this process, for each event class we end up with a list of concept scores. The event detector is then defined as the set of the top-$K$ scores and the corresponding concept labels. An example of such an event detector is given in Fig.~\ref{fig:E021_Detector} for the event class \textit{Attempting a bike trick}, where we observe that the three most dominant concepts are the following:\textit{ crash helmet}, \textit{bicycle-built-for-two}, and \textit{mountain bike}.

      \subsection{Applying the event detector to a video dataset}\label{subsec:apply_event_det}

	    The constructed event detector is used as shown in Fig.~\ref{fig:ZeroEx_Frameworks_2}. The DCNN-based detectors for the $N_c$ concepts in our concept pool are applied to the videos of the dataset, thus having these videos represented as vectors of concept detector output scores (hereafter called model vectors). The output scores that correspond to the $K$ concepts comprising the event detector are selected, and are compared with the corresponding values of the event detector. For this comparison, different choices of a distance function are possible, as shown in Table \ref{tbl:Design_Choices} (row C$4$); that is, we examine the use of following distance functions: Euclidean, Histogram Intersection, Chi-square, Kullback-Leibler, and cosine distance. For a chosen distance function, repeating the above process for all videos in a dataset we get a ranked list, in descending order, of the videos that most closely relate to the sought event.

      \subsection{Using multiple event detectors as pseudo-positive training samples}\label{subsec:training_ppos}

	    Figure~\ref{fig:ZeroEx_Frameworks_3} shows an alternative approach to using event detectors that are built as presented in section \ref{subsec:build_event_detector}. An event detector, i.e. a ranked list of concept scores (Fig.~\ref{fig:ZeroEx_Frameworks_1}, Fig.~\ref{fig:E021_Detector}) can be considered as a pseudo-positive sample for the respective event class. Since multiple event detectors can be obtained for a given event class by varying the design choices C$1$ to C$3$ of Fig.~\ref{fig:ZeroEx_Frameworks_1} (as shown in Table \ref{tbl:Design_Choices})), a set of pseudo-positive samples can be obtained for each event class. Negative samples for the given event class can also be obtained, in two ways. First, samples that are pseudo-positive for other event classes can be considered as pseudo-negative for a particular event class. Secondly, real videos can be selected from the Web, in analogy to how images are often selected as negative samples for training concept detectors from Web data, e.g. \cite{yan2003negative,li2013bootstrapping}. Using these pseudo-positives and (pseudo-)negatives, a new machine learning-based event detector can be obtained by training an SVM model (for instance using the RBF kernel). This can then be evaluated on any video dataset, following the application of trained concept detectors to the videos, (as in section \ref{subsec:apply_event_det}), to give us a new ranked list of event-related videos.


\section{Learning a video event detector from a few positive and related video examples}\label{sec:text_fewpos_comb}

      \subsection{Exploiting related videos for learning from a few examples}\label{subsec:fewpos_rel}
	    
	    As discussed in section \ref{sec:rel_work}, in the problem of video event detection, it is not unusual to be provided, or be able to find, some related video samples, i.e. videos that are closely related with a given complex event, but do not meet the exact requirements for being a true positive event instance. Exploiting related samples can be particularly interesting when only a few true positive samples are available, since in the opposite case, when an abundance of positive samples are available, one can effectively learn from them.
	    
	    In this section, Relevance Degree Support Vector Machine (RDSVM), proposed in \cite{tzelepis2013improving} for handling ``related'' training samples, is employed such that related samples are taken into consideration as weighted negative or weighted positive examples, where weighting is carried out completely automatically. RDSVM extends the standard SVM algorithm such that each training sample is assigned with a confidence value in $(0,1]$ indicating the degree of relevance of each training sample with the class it is related.
      
	    Let $\mathcal{X}=\{(\mathbf{x}_i, y_i)\colon \mathbf{x}_i\in\Bbb{R}^d, y_i\in\{0,\pm1\},i=1,\ldots,l\}$ be an annotated dataset of $l$ observations, where $\mathbf{x}_i\in\Bbb{R}^d$ is the feature vector representation of the $i$-th observation in the $d$-dimensional space with label $y_i\in\{0,\pm1\}$ denoting that the $i$-th observation is a positive ($y_i=+1$), a negative ($y_i=-1$), or a related 
	    ($y_i=0$) instance of the class. To allow the use of the RDSVM, the above is reformulated as $\mathcal{X}=\{(\mathbf{x}_i,y_i,u_i)\colon \mathbf{x}_i\in\Bbb{R}^d, y_i\in\{\pm1\},u_i\in\{0,1\},i=1,\ldots,l\}$, where $u_i$ is the so-called relevance label denoting that the $i$-th observation is a true ($u_i=0$) or a related ($u_i=1$) instance of the class $y_i$.
  
	    For the exploitation of the related observations, as proposed in \cite{tzelepis2013improving, wu2004incorporating}, each training sample 
	    $\mathbf{x}_i$ is associated with a adjustable confidence value $v_i$, and a monotonically increasing function $g(v_i)$ (called slack normalization function) is used to weight each slack variable $\xi_i$ denoting the loss introduced by a misclassified sample. In this way, support vectors (SVs) that are associated with a higher confidence value have greater contribution to the computation of 
	    the decision function. In \cite{tzelepis2013improving}, this function is modified so that only related class observations are associated with a confidence value $c_i\in(0,1]$ (called hereafter relevance degree) indicating the degree of relevance of the $i$-th observation with the class it is related. That is, $g$ is defined as follows:
	    \begin{equation}\label{eq:sec3eq1}
		  g(u_i) = 
			\left\{ 
			\begin{array}{ll}
			      1   & \mbox{if } u_i = 0, \\ 
			      c_i & \mbox{if } u_i = 1.
			\end{array} 
			\right. 
	    \end{equation}
	    Thus, the contribution of the related samples in the computation of the decision function can be regulated using appropriate relevance degrees $c_i$.
      
\begin{figure}
		  \centering
		  \includegraphics[width=9cm]{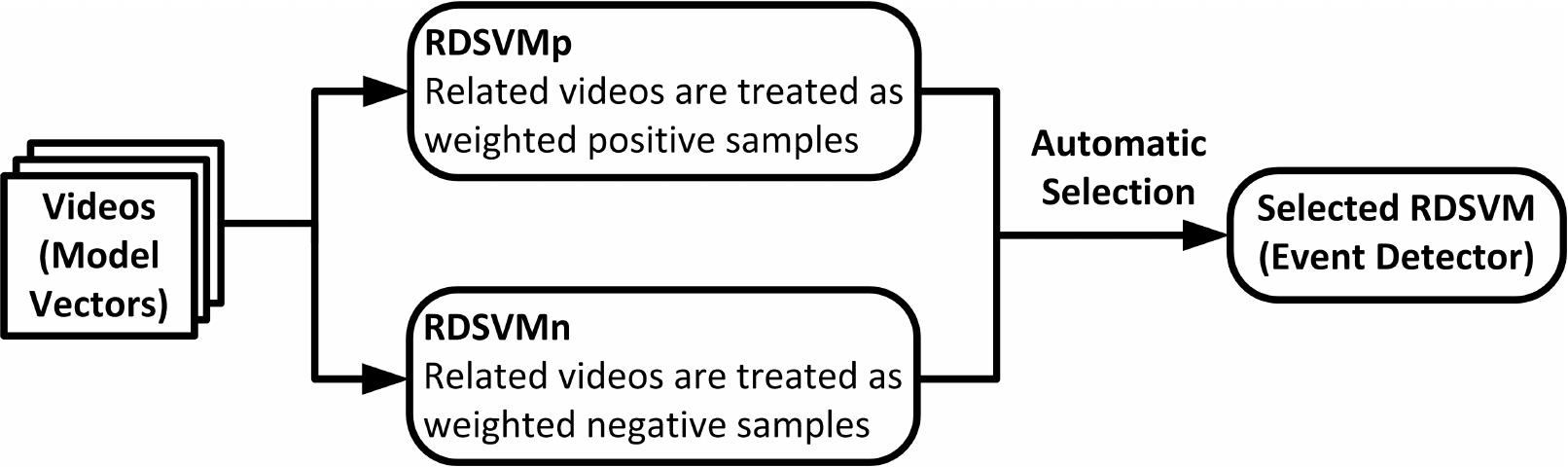}
		  \caption{RDSVM training framework.}
		  \label{fig:rdsvm_framework}
	    \end{figure}

	    In this study, we follow the approach proposed in \cite{tzelepis2013improving} for handling the related samples as a subclass of the positive or negative class, for which a global relevance degree, $c=c_i$ $\forall i$, is assigned to all related samples. Parameter $c$ is optimized using a cross-validation procedure. The proposed technique for the automatic selection of related samples as weighted positives, or weighted negatives, includes the following steps. First, an RDSVM is trained using the related videos as weighted positive examples (i.e., $y_i=+1$ and $u_i=1$). Then, another RDSVM is trained using the related videos weighted negative examples (i.e., $y_i=-1$ and $u_i=1$). The above classifiers are denoted in Fig.~\ref{fig:rdsvm_framework} as $\text{RDSVM}_p$ and $\text{RDSVM}_n$, respectively. 
	    
	    In both cases the basic parameters of the RDSVMs (e.g. $C$, $\gamma$ for an RDSVM with RBF kernel) are optimized by conducting cross-validation (grid search) with $c$ set to $1$, i.e. treating the related samples as pure positive and pure negative samples, respectively. Subsequently, the relevance degree parameter $c$ is optimized using cross-validation (line search) in the range $(0,1]$. After both $\text{RDSVM}_p$ and $\text{RDSVM}_n$ are trained in this way, one of the two is chosen (i.e., a parameter set $\{C,\gamma,c\}$, where $c\in[-1,1]$ is chosen) by looking at the average performance measure (e.g. average precision (AP)) values attained during cross-validation by the $\text{RDSVM}_p$ and $\text{RDSVM}_n$ (the one with the highest AP is selected). The chosen RDSVM is a learned event detector, that can be applied to a set of videos (where again the     video representation $\mathbf{x}_i$ can be the same as in section \ref{sec:textual}: a model vector produced by application of DCNN-based concept detectors).
	    
      \subsection{Treating pseudo-positive training samples as related ones}\label{subsec:fewpos_textual}
      
	    In section \ref{subsec:training_ppos} we discussed how we could use the output of processing an event's textual description as a set of pseudo-positive event samples, for training a standard SVM. When we also have a few true positive videos available for training, one possibility would be to merge the two positive/pseudo-positive sets and use their union for SVM training. Since, however, the pseudo-positive samples were artificially created and thus may be corrupt by errors/noise, a better option would be to treat them as ``related'' samples, rather than true positives. This can by straightforwardly achieved by use of the RDSVM methodology presented in the previous section. Evaluation of this approach will be presented in section \ref{subsec:rdsvm_exp}.


\section{Experimental results}\label{sec:exp_results}

      \subsection{Datasets and experimental setup}\label{subsec:datasets_exp_setup}
      
	    The proposed techniques are tested on the large-scale video dataset of the TRECVID Multimedia Event Detection (MED) 2014 task (hereafter referred to as MED$14$). The ground-truth annotated portion of it consists of three different video subsets: the ``pre-specified'' (PS) video subset ($2000$ videos, $80$ hours, $20$ event classes), the ``ad-hoc'' (AH) video subset ($1000$ videos, $40$ hours, $10$ event classes), and the ``background'' (BG) video subset ($5000$ videos, $200$ hours). Each video in the above dataset belongs to either one of $30$ target event classes (Table \ref{tab:med14_events}), or (in the case of the BG subset) to the ``rest of the world'' class . The above video dataset (PS+AH+BG) is partitioned such that a training and an evaluation set are created, as follows:
	    \begin{itemize}
		\item \textbf{Training Set}
		    \begin{itemize}
			\item $50$ positive samples per event class
			\item $25$ related (near-miss) samples per event class 
			\item $2496$ background samples (negative for all event classes)
		    \end{itemize}
		\item \textbf{Evaluation Set}
		    \begin{itemize}
			\item $\sim50$ positive samples per event class
			\item $\sim25$ related (near-miss) samples per event class
			\item $2496$ background samples (negative for all event classes)
		    \end{itemize}
	    \end{itemize}
	    For video representation, approximately $2$ keyframes per second were extracted. Each keyframe was represented using the last hidden layer of a pre-trained Deep CNN. More specifically, a $16$-layer pre-trained deep ConvNet network provided in \cite{simonyan2014very} was used. This network had been trained on the ImageNet data \cite{deng2009imagenet} and provides scores for $1000$ concepts. Thus, each keyframe has a $1000$-element vector representation. Then, a video-level model vector for each video is computed by taking the average of the corresponding keyframe-level representations.

	    \begin{table}[]
	    \scriptsize
	    \centering
	    \begin{tabular}{|l|l|}
	    \hline
	    \multicolumn{2}{|c|}{{\bf Target Events}}                                                     \\ \hline\hline
	    E021 - Attempting a bike trick          	&E036 - Felling a tree                         \\
	    E022 - Cleaning an appliance	            &E037 -  Parking a vehicle                      \\
	    E023 - Dog show                                &E038 -  Playing fetch                          \\
	    E024 - Giving directions to a location         &E039 -  Tailgating                             \\
	    E025 - Marriage proposal                       &E040 -  Tuning a musical instrument            \\
	    E026 - Renovating a home                       &E041 -  Baby Shower                            \\
	    E027 - Rock climbing                           &E042 -  Building a Fire                        \\
	    E028 - Town hall meeting                       &E043 -  Busking                                \\
	    E029 - Winning a race without a vehicle        &E044 -  Decorating for a Celebration           \\
	    E030 - Working on a metal crafts project       &E045 -  Extinguishing a Fire                   \\
	    E031 - Beekeeping                              &E046 -  Making a Purchase                      \\
	    E032 - Wedding shower                          &E047 -  Modeling                               \\
	    E033 - Non-motorized vehicle repair            &E048 -  Doing a Magic Trick                    \\
	    E034 - Fixing musical instrument               &E049 -  Putting on Additional Apparel          \\
	    E035 - Horse riding competition                &E050 -  Teaching Dance Choreography            \\ \hline
	    \end{tabular}
	    \caption{Target events in the employed TRECVID MED dataset.}
	    \label{tab:med14_events}
	    \end{table}  

      \subsection{Video event detection using the event's textual description}\label{subsec:text_desc}
      
	    The framework proposed in Section \ref{sec:textual} allows for different design choices in its various stages (Table \ref{tbl:Design_Choices}, and shaded processing stages in Fig.~\ref{fig:ZeroEx_Frameworks_All}). There are $5$ stages that can be parameterized: the ELM and CLM information sources selection (C$1$, C$2$a), the CLM textual vector weighting strategy (C$2$b), the matrix operator used (C$3$), and the distance function selected (C$4$). Based on the possible choices listed in Table \ref{tbl:Design_Choices}, $450$ different combinations are possible and were tested.
	    
	    Table \ref{tbl:Top10} presents the $10$ best-performing combinations in terms of mean average precision (MAP) across all $30$ events. Note that, when the ``Title'' is used for both ELM and CLM construction (processing stages C1 and C2a), meaning that both ELM and CLMs are represented by a single word or phrase ($N=M=1$), then all matrix operations result in the same score, making no difference in the final result. This is the reason why a dash sometimes appears in the C$3$ column of Table \ref{tbl:Top10}. Moreover, when ``Title'' is selected for the construction of the CLMs (C2a processing stage), there is no need for Bag-of-Words encoding, thus no use of weighting technique (Tf-Idf), since no enrichment by searching in Google or in Wikipedia was carried out. This is why there is no choice in the weighting stage (C2b) when ``Title'' is selected. As can be seen, the enrichment of the CLM through Google search, without Tf-Idf weighting and the usage of as much as possible information for constructing the EML resulted in the best result overall.
	    
	    In order to explore the effectiveness of the different design choice combinations, we experimented with changing one design choice at a time, keeping the choices for all other processing stages unaffected (as in the best-performing combination of Table \ref{tbl:Top10}). Table \ref{tbl:ELM} shows the performance of different ELMs in stage C$1$, clearly suggesting that the exploitation of as much as possible information for the particular parameter leads to better detection results. In Table \ref{tbl:CLM}, it is shown that 
	    enrichment of the CLM in stage C$2$a by searching in Google is the most effective way to do so. As Table \ref{tbl:Weght} shows, using Tf-Idf weighting does not seem to provide any improvement to the detection performance. Table \ref{tbl:Norm} suggests that the use of the Hausdorff distance, as a similarity matrix operator, outperforms the rest of the respective choices. Finally, Table \ref{tbl:Distances} shows the performance of different similarity measures (stage C$4$) in the event detection step (Fig.~\ref{fig:ZeroEx_Frameworks_2}), where the cosine and the Histogram Intersection measures outperform the rest of them.
     
	    In Table \ref{tbl:SVM_MAP}, we present the performance of the training step of our framework, as illustrated in Fig.~\ref{fig:ZeroEx_Frameworks_3} (denoted as $T_{10}$), compared to the best combination of the event detection step, shown in Fig.~\ref{fig:ZeroEx_Frameworks_2} (denoted as $T_{0}$). As discussed above, for $T_{10}$ we consider the event detectors generated as shown in Fig.~\ref{fig:ZeroEx_Frameworks_1} as pseudo-positive instances of the respective event class. Also, as discussed previously, negative samples can be obtained using two different approaches. First, the pseudo-positive samples from the rest of the event classes are used as pseudo-negative samples, and, secondly, real negative videos for all event classes, belonging to the ``background'' (BG) training subset, are used. Finally, a linear late fusion approach, using the arithmetic mean operator, is used in order to improve the individual detection results, denoted as $T_{0} \oplus T_{10}$. Hereafter, we will denote with $\oplus$ the late fusion scheme where the arithmetic mean operator is used for combining the event detection output scores. We can observe that the combination of the event detection stage, $T_0$, (Fig.~\ref{fig:ZeroEx_Frameworks_2}) with SVM training with pseudo-negative samples, $T_{10}$, (Fig.~\ref{fig:ZeroEx_Frameworks_3}) achieves the best performance, resulting in MAP equal to $12.38\%$.

	    \begin{table*}[!t]
		\centering
		\begin{tabular}{|ccccc||c|}
		    \hline
			 \textbf{C1}   & \textbf{C2a} & \textbf{C2b} & \textbf{C3}        & \textbf{C4}   &              \\ 
			 (ELM)         & (CLM)        & (Weighting)  & (Matrix Operation) & (Distance)    & \textbf{MAP} \\ \hline
			 AudioVisual   & Google       & No Tf-Idf    & Hausdorff          & Cosine        & 0.1111       \\
			 AudioVisual   & Google       & No Tf-Idf    & Hausdorff          & Histog\_Inter & 0.1109       \\
			 AudioVisual   & Google       & No Tf-Idf    & Hausdorff          & Kullback      & 0.1054       \\
			 Title         & Title        &  -           &   -                & Histog\_Inter & 0.1045       \\
			 Visual        & Google       & No Tf-Idf    & Hausdorff          & Cosine        & 0.1005       \\
			 Visual        & Google       & No Tf-Idf    & Hausdorff          & Histog\_Inter & 0.0991       \\
			 Title         & Title        &  -           &  -                 & Cosine        & 0.0988       \\
			 AudioVisual   & Google       & Tf-Idf       & Hausdorff          & Histog\_Inter & 0.0978       \\
			 Visual        & Google       & No Tf-Idf    & Hausdorff          & Kullback      & 0.0956       \\
			 AudioVisual   & Google       & Tf-Idf       & Hausdorff          & Cosine        & 0.0933       \\
		    \hline
		\end{tabular}
		\caption{Top-$10$ parameter combinations in terms of MAP.}
		\label{tbl:Top10}
	    \end{table*}

	    \begin{table*}[!t]
	    \centering
		  \begin{tabular}{|ccccc||c|}
		  \hline
			\textbf{C1} & \textbf{C2a}              & \textbf{C2b}               & \textbf{C3}                & \textbf{C4}             &              \\ 
			(ELM)       &     (CLM)                 & (Weighting)                & (Matrix Operation)         & (Distance)              & \textbf{MAP} \\ \hline
			AudioVisual & \multirow{3}{*}{Google}   & \multirow{3}{*}{No Tf-Idf} & \multirow{3}{*}{Hausdorff} & \multirow{3}{*}{Cosine} & 0.1111       \\
			Visual      &                           &         		     &                            &                         & 0.1005       \\
			Title       &                           &               	     &                            &                         & 0.0760       \\ \hline
		  \end{tabular}
		  \caption{Comparison of different Event Language Models (ELMs).}
		  \label{tbl:ELM}
	    \end{table*}
	      
	    \begin{table*}[!t]
	    \centering
		  \begin{tabular}{|ccccc||c|}
		  \hline
			 \textbf{C1}                            & \textbf{C2a} & \textbf{C2b}               & \textbf{C3}                & \textbf{C4}             &                        \\ 
			 (ELM)                                  & (CLM)        & (Weighting)                & (Matrix Operation)         &  (Distance)             & \textbf{MAP}           \\ \hline
			 \multirow{3}{*}{AudioVisual} 		& Google       & \multirow{3}{*}{No Tf-Idf} & \multirow{3}{*}{Hausdorff} & \multirow{3}{*}{Cosine} & 0.1111                 \\
								& Wikipedia    &                            &       						   &              & 0.0850  \\
								& Title	       &                            &           					   &              & 0.0760  \\ \hline 
		  \end{tabular}
		  \caption{Comparison of different Concept Language Models (CLMs).}
		  \label{tbl:CLM}
	    \end{table*}

	    \begin{table*}[!t]
	    \centering
		    \begin{tabular}{|ccccc||c|}
		    \hline
			\textbf{C1}                     & \textbf{C2a}            & \textbf{C2b} & \textbf{C3}                & \textbf{C4}             &              \\ 
			(ELM)                           & (CLM)                   & (Weighting)  & (Matrix Operation)         &  (Distance)             & \textbf{MAP} \\ \hline
			\multirow{2}{*}{AudioVisual} 	& \multirow{2}{*}{Google} & No Tf-Idf 	 & \multirow{2}{*}{Hausdorff} & \multirow{2}{*}{Cosine} & 0.1111       \\
		                                        &      			  & Tf-Idf   	 &   			      &                       	& 0.1005       \\ \hline
		    \end{tabular}
		    \caption{Comparison of different Weighting schemes.}
		    \label{tbl:Weght}
	      \end{table*}
	      
	      \begin{table*}[!t]
		    \centering
		    \begin{tabular}{|ccccc||c|}
		    \hline
			\textbf{C1}                  & \textbf{C2a}             & \textbf{C2b}               & \textbf{C3}        & \textbf{C4}              &              \\ 
			(ELM)                        & (CLM)                    & (Weighting)                & (Matrix Operation) &  (Distance)             & \textbf{MAP}  \\ \hline
			\multirow{5}{*}{AudioVisual} & \multirow{5}{*}{Google}  & \multirow{5}{*}{No Tf-Idf} & Hausdorff	  & \multirow{5}{*}{Cosine} & 0.1111        \\
				   		     &                          &                            & $l_2$              &                         & 0.0833        \\
						     &                          &                            & Frobenius          &                         & 0.0827	    \\
						     &                          &                            & $l_{\infty}$       &                         & 0.0828 	    \\
						     &                          &                            & Max	          &                         & 0.0804 	    \\ \hline
			\end{tabular}
		    \caption{Comparison of various norms of the $N\times M$ distance matrix.}
		    \label{tbl:Norm}
	      \end{table*}
	      	      
	    \begin{table*}[!t]
	    \centering
		    \begin{tabular}{|ccccc||c|}
		    \hline
			 \textbf{ELM}                 & \textbf{CLM}            & \textbf{Weighting}         & \textbf{Matrix Operation}  & \textbf{Distance} & \textbf{MAP} \\ \hline
			 \multirow{5}{*}{AudioVisual} & \multirow{5}{*}{Google} & \multirow{5}{*}{No Tf-Idf} & \multirow{5}{*}{Hausdorff} & Cosine            & 0.1111       \\
					              &                         &                            &                            & Histog\_Inter     & 0.1109       \\
					              &                         &                            &                            & Kullback          & 0.1054       \\
					              &                         &                            &                            & $X^2$             & 0.0832       \\
					              &                         &                            &                            & Euclidean         & 0.0690       \\ \hline
		    \end{tabular}
	      \caption{Comparison of different similarities measures.}
	      \label{tbl:Distances}
	      \end{table*}

	      \begin{table}[h]
		    \scriptsize
		    \centering
		    \begin{tabular}{|c|cccc|}
		    \hline
			  \begin{tabular}[c]{@{}c@{}}\textbf{Experimental}\\ \textbf{Scenario}\end{tabular}  & $\mathbf{T_{0}}$    & \begin{tabular}[c]{@{}c@{}}$\mathbf{T_{10}}$\\ \textbf{Pseudo-Negatives}\end{tabular} & \begin{tabular}[c]{@{}c@{}}$\mathbf{T_{10}}$\\ \textbf{Real Negatives}\end{tabular} & \begin{tabular}[c]{@{}c@{}} $\mathbf{T_{0} \oplus T_{10}}$ \end{tabular} \\ \hline\hline
			  MAP                                                                                & $0.1111$     & $0.10125$                                                                          & $0.0594$                                                                           & $\mathbf{0.1238}$                                                         \\ \hline
		    \end{tabular}
		    \caption{The performance of the developed techniques in AP, and across $30$ events in MAP.}
		    \label{tbl:SVM_MAP}
	      \end{table}

	      We compare the proposed method with the E-Lamp framework, a state-of-the-art system that participated in the TRECVID 2014 MED task~\cite{yuinformedia} and is described in detail in ~\cite{jiang2015bridging}. As mentioned in section \ref{subsec:rel_work_zero}, the E-Lamp system consists of four major subsystems, namely Video Semantic Indexing (VSIN), Semantic Query Generator (SQG), Multimodal Search (MS) and Pseudo-Relevance Feedback (PRF). The VSIN subsystem represents the input videos as a set of low- and high-level features from several modalities. The high-level features, i.e. the result of semantic concept detection, are used as input to the SQG subsystem, in which the textual description of an event class is translated into a set of relevant concepts termed \textit{system query}. The system query is then used in the MS subsystem as input to several well-known text retrieval models in order to find the most relevant videos. These results can be then refined by the PRF subsystem.

	      As SQG leads to the creation of an event detector using semantic concepts, a correspondence exists (and comparison is possible) with our approach to build an event detector as described in sections \ref{subsec:Lang_Models} and \ref{subsec:build_event_detector} (Fig.~\ref{fig:ZeroEx_Frameworks_1}). Similarly, the MS subsystem corresponds to (and can be compared with) our event detection module presented in section \ref{subsec:apply_event_det} (Fig. \ref{fig:ZeroEx_Frameworks_2}). We compared with four SQG approaches that are presented in the E-Lamp system. These are: i) Exact word matching, ii) WordNet mapping using Wu \& Palmer measure (Wu) iii) WordNet mapping using the structural depth distance in WordNet hierarchy (Path), and iv) Word Embedding Mapping (WEP). Concerning the MS stage, we compared with the following retrieval methods: the Vector Space Model~\cite{Younessian}, the Okapi BM25, and two unigram language models with Jelinek-Mercer smoothing (LM-JM) and Dirichlet smoothing (LM-DL) respectively~\cite{zhai2004study}.

	      Table~\ref{tbl:CMU_CERTH} shows the performance, in terms of mean average precision (MAP), of the above combinations in comparison to the best-performing event detector and distances proposed in the present work (Table \ref{tbl:Distances}). From this Table it is clear that the proposed method for building an event detector outperforms the rest of the compared methods, irrespective of the similarity measure that they are combined with. Out of the event detection creation methods of~\cite{jiang2015bridging}, the \textit{exact word} seems to perform considerably better than the others (but much worse than the proposed method). This is because the concept labels from our concept pool that are most related to an event are often well-represented in the event's textual description, e.g. for the event \textit{Beekeeping}, the word bee is observed 31 times, and this word is directly associated with the concepts \textit{bee} and \textit{bee eater}. The WordNet and WEP mappings on the other hand, are not always successful in finding the semantic similarity between two words. Regarding the compared similarity measures, the VSM and LM-DL generally perform better than BM25 and LM-JM, but the proposed cosine and Histogram Intersection distances are consistently among the top-performing measures. It should be noted that the number of visual concepts used in~\cite{jiang2015bridging} is significantly greater than the 1000 concepts used throughout our experiments, and other modalities (e.g. audio) are also exploited in~\cite{jiang2015bridging}; this explains the often higher MAP values that are reported in the latter work.

	      \begin{table*}[!t]
	      \centering
	      \begin{tabular}{|c||cccc|cc|}
	      \hline
	      
	      \multirow{3}{*}{\begin{tabular}[c]{@{}c@{}}Event detector\\creation\end{tabular}} & \multicolumn{4}{c|}{Similarity measures~\cite{jiang2015bridging}}  & \multicolumn{2}{c|}{\begin{tabular}[c]{@{}c@{}}Similarity measures \\ (proposed)\end{tabular}} \\ \cline{2-7} 
				      & {\bf VSM} & {\bf BM25} & {\bf LM-JM} & {\bf LM-DL} & {\bf Cosine}      & {\bf \begin{tabular}[c]{@{}c@{}}Histogram\\ Intersection\end{tabular}}     \\ \hline
	      {\bf WordNet - Wu}      & 0.0205	& 0.0201	& 0.0250	& 0.0319	& 0.0222	& 0.0318  \\
	      {\bf WordNet - Path}    & 0.0333	& 0.0221	& 0.0310	& 0.0379 	& 0.0359 	& 0.0434	 \\
	      {\bf Exact word}        & 0.0833	& 0.0287  	& 0.0541	& 0.0568	& 0.0828	& 0.0801	 \\
	      {\bf WEM}               & 0.0429    & 0.0232    & 0.0269	& 0.0331	& 0.0427	& 0.0418 \\ \hline
	      {\bf Proposed}          & 0.0912    & 0.0980   	& 0.0392	& 0.0993	& \textbf{0.1111}	& 0.1109 \\ \hline
	      \end{tabular}
	      \caption{Comparison between proposed and compared methods.}
	      \label{tbl:CMU_CERTH}
	      \end{table*}

      \subsection{Learning from a few positive and related video examples}\label{subsec:few_rel}
      
	    In this section, we validate the performance of the proposed framework for handling related samples for the problem of learning video event detectors from a few positive samples. To this end, we compare the following approaches in order to investigate whether and in what way it is beneficial to use related samples in training: a) using no related samples, b) related samples are used as pure negative ones, as in \cite{douze2014inria}, c) related samples are used as pure positive ones, again as in \cite{douze2014inria}, and d) related samples are used as weighted negative or positive ones under the RD-SVM framework of Section \ref{subsec:fewpos_rel}, which involves an automatic procedure for selecting both the labels of the related samples and their weights.
      
	    As discussed in section \ref{subsec:datasets_exp_setup}, the MED$14$ dataset provides $50$ positive samples per each of the $30$ event classes. However, we want to simulate the case where only a few positive samples are available for training, thus we choose to use only $10$ training samples per event class. To this end, for each experiment we randomly draw a subset of $10$ positive samples for each event class, and we repeat this $10$ times. That is, for each compared approach in this section, the obtained performance of the corresponding classifier (RD-SVM or standard SVM) is averaged over $10$ iterations. For the approaches that use related samples, $10$ such samples were chosen from the pool of $25$ related samples that are available for each event class; these, as suggested in \cite{tzelepis2013improving}, were selected as the $10$ that are the nearest to the median of all $25$ related samples in the employed feature space.
      
	    Table \ref{tab:rdsvm} shows the results of these comparisons, in terms of mean average precision (MAP), across the $30$ event classes of the MED$14$ dataset. We observe that when related samples are used as pure negative or pure positive ones as in \cite{douze2014inria}, the overall detection performance is lower than the baseline approach which does not use these samples at all ($P_{10}$). In contrast to this, the proposed approach that treats related samples as positive/negative ones with automatic weighting achieved better performance than the baseline, reaching a MAP of $18.95\%$.
	    
\begin{figure*}[t]
		\centering
		    \begin{subfigure}[b]{\linewidth}
			    \centering
			    \includegraphics[width=15cm]{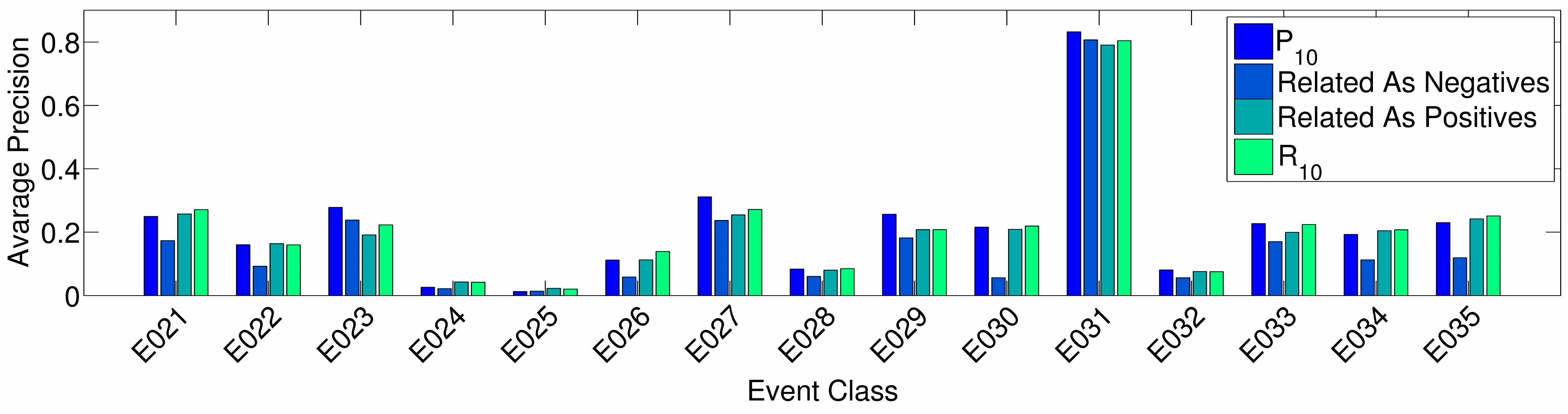}
			    \caption{}
			    \label{subfig:E021_E035}
		    \end{subfigure}
		    \\
		    \begin{subfigure}[b]{\linewidth}
			    \centering
			    \includegraphics[width=15cm]{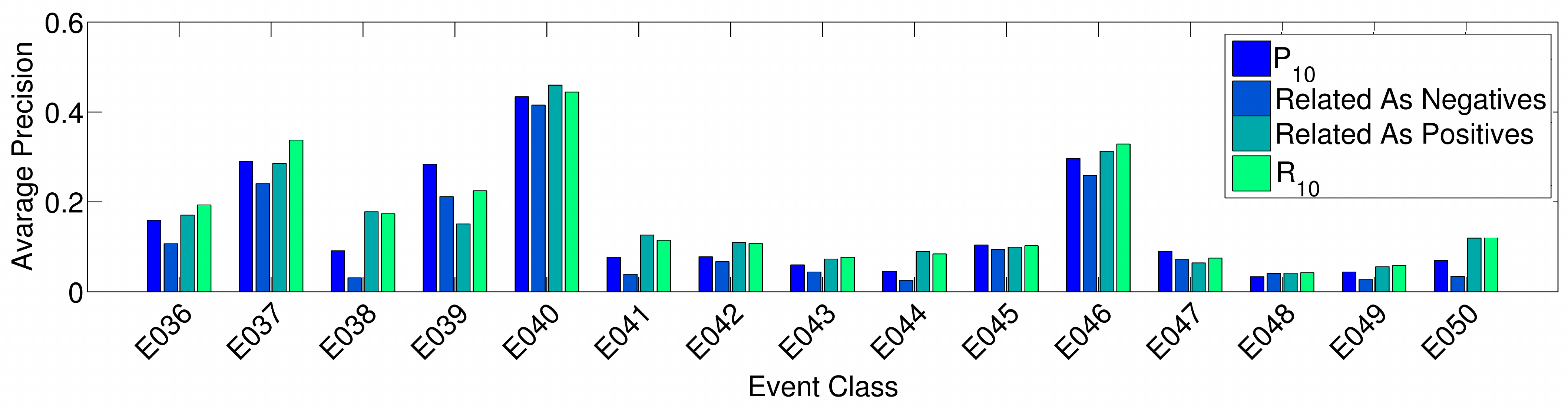}
			    \caption{}
			    \label{subfig:E036_E050}
		    \end{subfigure}
		  \caption{Experimental evaluation of various methods for treating related samples for event classes (a) E021-E035 and (b) E036-E050.}
		  \label{fig:rdksvm_comp}
	    \end{figure*}

	    In Fig.~\ref{fig:rdksvm_comp} the results of the above comparisons are given separately for each of the $30$ event classes of the dataset. As can be seen, despite the fact that treating related samples as either pure negatives or pure positives jointly for all the event classes leads to worse average detection performance (MAP), compared to the case where they are not used at all in the training process, there are event classes where it is beneficial to use them in such a way. Specifically, for $15$ out of $30$ events it is better to use related samples as pure positives rather that to exclude them from the training process, and similarly for one event it is beneficial to use them as pure negatives. Automatically selecting to use related samples as weighted negatives or weighted positives using RD-SVM \cite{tzelepis2013improving}, on the other hand, is better than excluding them from the annotation process for $19$ out of the $30$ event classes. Moreover, the automatic weight selection leads to better results than both approaches that use related samples without any weight for $18$ out of the $30$ events. The above confirm our hypothesis that treating related samples as weighted negatives/positives using RD-SVM \cite{tzelepis2013improving} can lead to better event detection performance.

	    \begin{table}[h]
	    \footnotesize
	    \centering
		  \begin{tabular}{|c|cccc|}
		  \hline
		  {\bf \begin{tabular}[c]{@{}c@{}}Experimental\\ Scenario\end{tabular}} & {\bf \begin{tabular}[c]{@{}c@{}}$\mathbf{P_{10}}$\\ (baseline)\end{tabular}} & {\bf \begin{tabular}[c]{@{}c@{}}Related as \\ negatives\\ (as in \cite{douze2014inria})\end{tabular}} & {\bf \begin{tabular}[c]{@{}c@{}}Related as \\ positives\\ (as in \cite{douze2014inria})\end{tabular}} & {\bf \begin{tabular}[c]{@{}c@{}}$\mathbf{R_{10}}$\\ (proposed)\end{tabular}} \\ \hline\hline
		  {\bf MAP}                                                             & $0.1808$                                                              & $0.1368$                                                                                 & $0.1796$                                                                                 & $\mathbf{0.1895}$                                                              \\ \hline
		  \end{tabular}
		  \caption{Comparisons of approaches using related samples for learning an event detector.}
		  \label{tab:rdsvm}
	    \end{table}

      \subsection{Combining textual event detectors with learning from a few positive and related samples}\label{subsec:rdsvm_exp}
	    
	    Our first attempt to combine textual event detectors and learning from a few examples is based on exploiting the pseudo-positive samples, which were computed according to section \ref{subsec:training_ppos}, as related samples in RD-SVM. We chose a subset of $10$ pseudo-positive samples for each event class, using a same selection strategy as in the case of related samples in section \ref{subsec:few_rel}; that is, the $10$ nearest to the median pseudo-positive sample for each event class were selected. Using RDSVM for handling pseudo-positive samples as weighted negative or positive ones (with automatic weighting selection) resulted in a MAP equal to $18.26\%$, across $30$ events (denoted as $R_{10p}$ in Table \ref{tab:rdsvm_fusion}). It is worth noting that, similarly to the previous results, this experimental result stands for the average of $10$ iterations (using $10$ different, randomly selected sets of $10$ positive examples each). We observe that, using pseudo-positive samples this way outperforms learning from solely $10$ positive samples by a small margin.
	    
	    In a second attempt to further combine text-based and learning-based event detectors, we examine the late fusion of detectors. As shown in the last column of Table \ref{tbl:SVM_MAP} , $T_{0} \oplus T_{10}$ denotes the best approach for learning video event detectors using the textual description of each event class alone, resulting in a MAP equal to $12.38\%$. The results of combining the above approach, as well as $R_{10p}$ (which already jointly uses textual information and real training samples) with the $P_{10}$ and $R_{10}$ learned detectors (Table \ref{tab:rdsvm}), are shown in Table \ref{tab:rdsvm_fusion}. As we can see, the best-performing combination is that of $R_{10}$ and $R_{10p}$, which achieves MAP equal to $20.11\%$. This is higher that the best performance that is achieved in our experiments using visual examples alone ($R_{10}$), and much higher than that achieved using only textual information about the sought event, highlighting the importance of combining visual examples and a textual description of the event for learning. Other combinations, most notably that of $T_{0} \oplus T_{10}$ with $P_{10}$, $R_{10}$ or $R_{10p}$, do not seem to offer any improvement over $R_{10p}$ alone. This can be attributed to the fact that the $T_{0} \oplus T_{10}$ detector is much weaker than $P_{10}$, $R_{10}$, and $R_{10p}$, thus introducing mostly noise to the results of the latter at the late fusion stage.

	    \begin{table}[h]
		  \footnotesize
		  \centering
		  \begin{tabular}{|lc|}
		  \hline
		  \textbf{Combinations of event detectors}   & \textbf{MAP}        \\ \hline\hline
		  $R_{10p}$                                  & $0.1826$            \\ \hline
		  $T_{0}  \oplus T_{10} \oplus P_{10}$       & $0.1743$            \\
		  $T_{0}  \oplus T_{10} \oplus R_{10}$       & $0.1662$            \\
		  $T_{0}  \oplus T_{10} \oplus R_{10p}$      & $0.1647$            \\ \hline
		  $P_{10} \oplus R_{10}$                     & $0.1893$            \\
		  $P_{10} \oplus R_{10p}$                    & $0.1971$            \\
		  $P_{10} \oplus R_{10} \oplus R_{10p}$      & $0.1965$            \\ \hline
		  $R_{10} \oplus R_{10p}$                    & $\mathbf{0.2011}$   \\ \hline
		  \end{tabular}
		  \caption{Various combinations of textual event detectors with learning from few positive, related, and pseudo-positive training samples.}
		  \label{tab:rdsvm_fusion}
	    \end{table}
	    
 	    \begin{figure*}
 		  \centering
 		  \includegraphics[width=15cm]{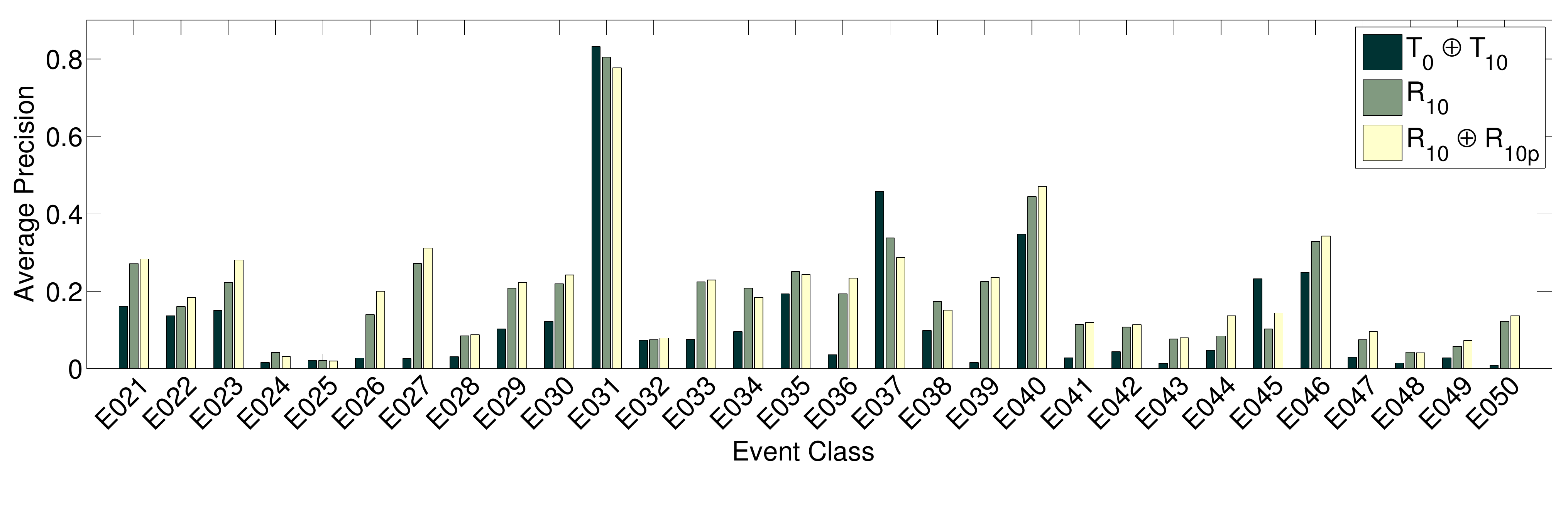}
 		  \caption{Event detection performance (AP) for $30$ event classes of MED14 dataset using three different fusion approaches.}
 		  \label{fig:Fusion_Hist}
 	    \end{figure*}
 	    
 	    In Fig.~\ref{fig:Fusion_Hist}, we present the event detection results per event class, using Average Precision (AP), for i) the best-performing proposed approach that learns from the events' textual descriptions (last column of Table \ref{tbl:SVM_MAP}); ii) the best-performing proposed approach for learning from a few positive and related examples (last column of Table \ref{tab:rdsvm}); and, iii) the best combination of Table \ref{tab:rdsvm_fusion} (last row) for exploiting both video examples and a textual description of the event. We observe that the latter combination outperforms the former two approaches for $22$ of the $30$ event classes. Notable exceptions, where using just the textual description of the event performs the best, are events E031 (\textit{Beekeeping}), E037 (\textit{Parking a vehicle}), and E045 (\textit{Extinguishing a Fire}). This can be attributed to the fact that we have in our concept pool a wealth of concepts related to these events (e.g. for E037: \textit{beach wagon, car mirror, electric locomotive, minibus, parking meter, recreational vehicle, sports car, streetcar, etc.}; for E045: \textit{fire engine, fire screen, fireboat, gas pump, stove, water tower, etc.}) and at the same time the textual description which is performed for these events allows us to identify those related concepts (e.g. for E031, the top-$4$ concepts that $T_{0} \oplus T_{10}$ identifies are: \textit{honeycomb}, \textit{apiary}, \textit{bee}, \textit{bee eater}). In contrast to this, for several of other events only one or two concept (or even no concepts at all) closely relate to them, e.g. for event E047 (\textit{Modeling}) the closest-related concept that is included in our concept pool is \textit{kimono} and, similarly, concept \textit{grocery store} for event E046 \textit{Making a Purchase}.


\section{Conclusion}\label{sec:conclusion}
      
      In this paper we proposed a framework for learning video event detectors from solely a textual description of an event class, or from a very few positive and related training samples. We identified a general learning framework and studied the impact of various design choices for different stages of this framework. For exploiting related video samples we employed an SVM extension (RDSVM) such that related samples are automatically treated as weighted negative or positive samples. The experimental evaluation of the proposed approaches, as well as the 
      combination of them on the challenging, large-scale TRECVID MED $2014$ video dataset verified the applicability of the proposed methods in the cases where positive samples are not available, or scarce, and provided useful insight on how to train video event detectors under such conditions.


\section{Acknowledgment}\label{sec:acknowledgment}
      
      This work was supported by the European Commission under contracts FP7-600826 ForgetIT and FP7-287911 LinkedTV.

\bibliographystyle{IEEEtran}
\bibliography{bibliography}

\end{document}